\documentclass[nohyperref]{article}
\usepackage{celyn}
\usepackage[accepted]{icml2023}

\theoremstyle{plain}

\theoremstyle{definition}

\theoremstyle{remark}

\begin{acronym}[EDeNN] 
	\acro{EDeNN}{Event Decay Neural Network}
	\acrodefplural{EDeNN}{Event Decay Neural Networks}
	\acro{EDeC}{Event Decay Convolution}
	\acro{RL}{Reinforcement Learning}
	\acro{PPO}{Proximal Policy Optimisation}
	\acro{CNN}{Convolutional Neural Network}
	\acro{SNN}{Spiking Neural Network}
	\acrodefplural{SNN}{Spiking Neural Networks}
	\acro{CERiL}{Continuous Event-based Reinforcement Learning}
	\acro{eVAE}{event Variational Autoencoder}
	\acro{DQN}{Deep Q-Network}
	\acro{TRPO}{Trust Region Policy Optimization}
	\acro{ODE}{Ordinary Differential Equations}
\end{acronym}

\graphicspath{{images/},{diagrams/}}
\begin{document}

\twocolumn[
	\icmltitle{CERiL: Continuous Event-based Reinforcement Learning}
	\begin{icmlauthorlist}
		\icmlauthor{Celyn Walters}{cvssp}
		\icmlauthor{Simon Hadfield}{cvssp}
	\end{icmlauthorlist}
	\icmlaffiliation{cvssp}{Centre for Vision, Speech and Signal Processing (CVSSP), University of Surrey, Guildford, UK}
	\icmlcorrespondingauthor{Celyn Walters}{celyn.walters@surrey.ac.uk}
	\icmlcorrespondingauthor{Simon Hadfield}{s.hadfield@surrey.ac.uk}
	\icmlkeywords{Machine learning, Event camera, Reinforcement learning}
	\vskip 0.3in
]
\printAffiliationsAndNotice{}

\begin{abstract}\label{abstract}
This paper explores the potential of event cameras to enable continuous time \acl{RL}.
We formalise this problem where a continuous stream of unsynchronised observations is used to produce a corresponding stream of output actions for the environment.
This lack of synchronisation enables greatly enhanced reactivity.

We present a method to train on event streams derived from standard \acs{RL} environments, thereby solving the proposed continuous time \acs{RL} problem.
The \acs{CERiL} algorithm uses specialised network layers which operate directly on an event stream, rather than aggregating events into quantised image frames.

We show the advantages of event streams over less-frequent RGB images.
The proposed system outperforms networks typically used in \acs{RL}, even succeeding at tasks which cannot be solved traditionally.
We also demonstrate the value of our \acs{CERiL} approach over a standard \acs{SNN} baseline using event streams.
\end{abstract}

\section{Introduction}\label{sec:intro}
\begin{figure*}[t]\centering
	\includegraphics[width=\linewidth]{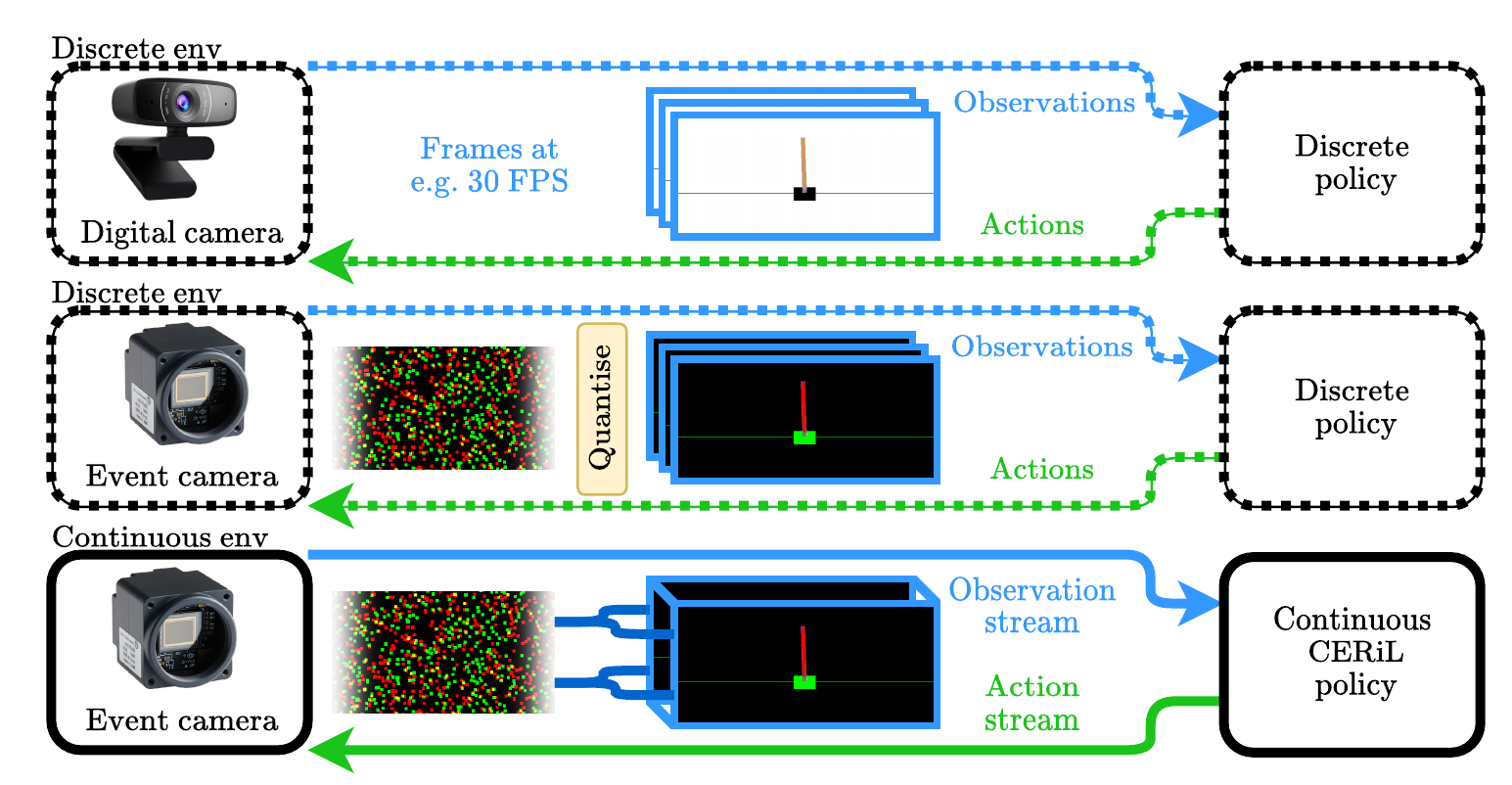}%
	\caption{\label{fig:front}
		A typical digital camera produces frames at a set rate, leading to a discrete policy and actions at regular time intervals as shown in row 1.
		Traditional event camera systems follow the same paradigm by aggregating the event stream as shown in row 2.
		Our proposed continuous \acs{CERiL} formulation shown in row 3 produces a continuous action stream directly from the observation streams.
	}
\end{figure*}
\ac{RL} is an approach to learn long term strategies to maximise a given reward signal with high generalisation to unseen environments.
This is generally framed as a trial-and-error type exploration problem, which attempts to generalise to unseen environmental states based on similar previously seen states.
This approach to learning has obvious parallels to learning in nature.

These parallels are even more apparent for \ac{RL} systems driven by visual input observations.
The majority of prior work rely on `ground truth' observations, \ie~the agent position and task-related information are provided directly.
In the real world though, it is often impossible to obtain this information and a visual stream of a much higher dimensional state representation is unavoidable.
Approaches that directly operate on raw image frames have not been as extensively studied.
Several algorithms have been proposed for simple, synthetic images and video game environments like Atari games~\cite{Kaiser2019}, but the high dimensional state space, containing potentially tens of thousands of pixels, makes the learning problem far more challenging.

One aspect of vision-based \ac{RL} (and indeed all \ac{RL}) which differs from nature is the treatment of time.
It is almost always assumed that time is split into a number of discrete and equally spaced intervals or `steps'.
Each step includes an observation of the environment, which leads to an action, which in turn updates the environmental state.
This makes sense for vision based systems using traditional frame based cameras, where observations of the environment generally constitute `frames' which are regularly spaced in time.
However, modern developments in sensor technology called `event cameras' may make it possible to rectify this mismatch.

Event cameras are a type of asynchronous visual sensor.
This means that every pixel triggers a signal independently based on its own detected brightness change.
There is no `shutter' mechanism to trigger a synchronised measurement from all pixels.
As a result the output of the camera is a continuous stream of event signals, with variable rate and interval depending on the scene contents.
This provides greatly reduced latency and environmental changes are recorded within microseconds.
In addition the sensor avoids motion blur, and provides greatly enhanced dynamic range and low power consumption.

This constant stream of observations is much closer to how a biological vision system operates.
However, without discrete `frames' it becomes more challenging to apply standard deep-learning techniques.
As a result, lots of work utilising event cameras for deep-learning has followed a similar approach.
The event stream is aggregated into a number of regularly spaced `event images' which capture information such as the number of events, or the variance in the timestamps.
The obvious downside to this approach is that many of the advantages that the sensor provided are lost.
In particular, the aggregation reintroduces a regular `shutter' which means that the system cannot react rapidly to its observations.

In contrast, this paper formalises the \ac{RL} problem in the continuous time domain, where a stream of input events leads to a continuous stream of output actions.
We then demonstrate how a continuous actor and critic network can be produced using a recently developed deep learning approach known as \acp{EDeNN}~\cite{Walters2022}.
This applies a unique form of 3D spatio-temporal convolution directly on the continuous observation volume.
The result is a continuous stream of actions and value estimates respectively.
The interaction with the environment and the update of the network parameters still happens at discrete times.
However, these times are irregularly spaced and can be far more frequent than the traditional environmental observation, limited only by the speed of the control loop.
Our approach is contrasted against the two alternatives in \cref{fig:front}.
To summarise, the contributions of this work are:
\begin{enumerate}
	\item A new formalisation of the continuous \ac{RL} problem, including definitions of the loss functions for \ac{PPO}.
	\item A framework to turn any OpenAI gym environment into a continuous \ac{RL} problem, as long as it implements a traditional render function.
	\item A novel framework based on the \ac{PPO} paradigm to solve these continuous \ac{RL} problems.
	We refer to this algorithm as \acl{CERiL}.
\end{enumerate}
The code supporting all these contributions will be made available to the community to help spark further research in this area.

\section{Related Work}\label{sec:lit}
\ac{RL} became mainstream when a system was demonstrated playing backgammon at an expert level~\cite{Tesauro1995}, and actually advanced understanding of the game's theory.
Since then, approaches have become more complex and capable.
In 2015, Mnih \etal~introduced \ac{DQN}, which was the first mainstream approach which operated on a visual render stream, and could play many Atari 2600 games at a human level~\cite{Mnih2015}.
\ac{DQN} is an off-policy algorithm, which is named as such because the policy used for behaviour is not necessarily the policy used to populate a replay buffer with experience.
More recently popular is \ac{PPO}~\cite{Schulman2017}, an on-policy algorithm which improves upon \ac{TRPO}~\cite{Schulman2015a} by requiring only first-order rather than second-order gradients.
\ac{PPO} makes use of a `Trust Region', and significant policy updates are `clipped' which mitigates the effect of adverse policy changes.

Due to large amount of agent-in-the-loop experience which needs to be collected for \ac{RL}, it is very common for a simulation to be used.
OpenAI gym makes effort to standardise environments for easier distribution~\cite{Brockman2016}.
It includes simple physics models such as CartPole and MountainCar, as well as providing access to Atari games.
Event camera research is an emerging field, and there are a few simulation tools available.
ESIM~\cite{Rebecq2018} is able to generate event streams from a variety of sources, from 2D images or 3D models.
It is purposed towards generating datasets, which are scarce because the cost of event cameras is prohibitive.
AirSim~\cite{Madaan2020} is a multi-sensor simulator for drones and cars, more directed at robotics simulation.
Vemprala \etal~make use of AirSim and present an \ac{RL} approach using \ac{PPO}~\cite{Vemprala2021}.
They propose an `\ac{eVAE}' to preserve temporal information, inspired by the growing popularity of Transformer networks~\cite{Vaswani2017}.
\acp{SNN} are a type of network which, like event cameras, are biologically inspired.
There have been a few attempts at using \acp{SNN} in an \ac{RL} framework, for example, for maze-solving~\cite{Florensa2017} and drone flight control~\cite{Zanatta2022}.
One limitation with \acp{SNN} is that a large time period is typically required for inference, a problem for time-limited tasks.

The most common \ac{RL} environments have discrete action spaces.
Many of the continuous-control use the MuJoCo physics engine~\cite{Todorov2012}.
These take a floating point representing an action as input, but they operate at a discrete rate.
There are some existing continuous-time \ac{RL} approaches, but many of those require a dedicated engine and a known dynamics model.

Yıldız \etal~point out a divergence between continuous-time and discrete-time trajectories in CartPole~\cite{Yildiz2021}.
Although simulating surrogate \ac{ODE} dynamics is highly accurate, it at least depends on the physics state of a given environment to be available.
Our framework has to ability to generate a continuous observation space from existing OpenAI Gym environment without requiring a specialised simulation.

\section{Methodology}\label{sec:methodology}
We will first describe our continuous time formulation of the \ac{RL} problem.
This ensures the algorithm is suitable for use with asynchronous, irregularly sampled state-spaces, such as those resulting from high speed event cameras.

In traditional \ac{RL}, times are sampled at equally spaced integer steps (\ie~\(t \in \mathbb{Z}\)).
A rollout (\(\Omega\)) is thus traditionally defined as an ordered set of states \(\Omega = \{s_0, s_1, \ldots, s_T\}\).
In contrast, in this work we define a continuous time domain (\(t \in \mathbb{R}\)) where a state may be sampled from the rollout volume at any time as
\begin{equation}
	s_t(x, y, c) = \Omega(x, y, c, t)
	,
\end{equation}
where \(x,y,c\) represent the spatial positions (\(x, y\)) and colour channels (\(c\)) of the resulting state.
It is worth noting that the rollout volume \(\Omega\) may be sparsely populated.
Regardless, the time domain is continuous and states may be sampled as densely as the control scheme can operate.
Indeed the sampling of the states may be reactionary and uneven.

The reward function in an \ac{RL} system traditionally provides a scalar reward signal for a given state-action pair.
We instead define a `reward density function' \(r(a_t, \Omega, t)\).
This is queried to return the reward density over time at a particular point in the rollout, given action \(a_t\).
The need to define this reward density function could potentially place some limitations on the environments which can be used.
However, in practice the reward schemes for most traditional \ac{RL} environments can be easily adapted to fit this definition.
For example, `keep alive' or `finish quickly' style environments like CartPole or MountainCar will respectively provide a constant positive or negative reward per timestep.
In these cases the reward density can be similarly defined as a constant function.
There also exist many `all-or-nothing' type sparse reward environments.
Here the reward is determined solely by the final state of the rollout.
In this case the reward density function is set to zero at all points other than a brief interval at the end of the rollout.
More generally, for any existing discrete environment we can define a valid reward density function.
The density function must only obey the constraint
\begin{equation}
	\hat{r}_\tau(a_\tau, s_\tau) = \int_{t = \tau - 1}^\tau r(a_t, \Omega, t)
	,
\end{equation}
where \(\hat{r}\) is the reward function of the original discrete time environment.

Given our definition of the reward density function, we can now define the return for a time segment of a rollout as
\begin{equation}
	R(\Omega, t_0, T) = \int_{t = t_0}^T \gamma^{t - t_0} r(a_t, \Omega, t)
	,
\end{equation}
where \(\gamma\) is the discount factor for future rewards.

We next define the value function for a state at a given time \(t\) within the rollout as \(V^\pi(\Omega, t) = R(\Omega, t, \infty)\).
As detailed later, this value function encodes the expected return for policy \(\pi(a | \Omega, t)\), starting at state \(s_t\) within rollout \(\Omega\).
Finally, this value function now enables us to define the continuous time variant of the advantage function~\cite{Schulman2015} for a particular action sequence \(\hat{a}\) as
\begin{equation}
	A(\Omega, \tau, \hat{a}) = \left(\int_{t = \tau}^\infty \gamma^{t - \tau} r(\hat{a}, \Omega, t) dt\right) - V^\pi(\Omega, t)
	.
\end{equation}

\begin{figure}[h]\centering
	\includegraphics[width=\linewidth]{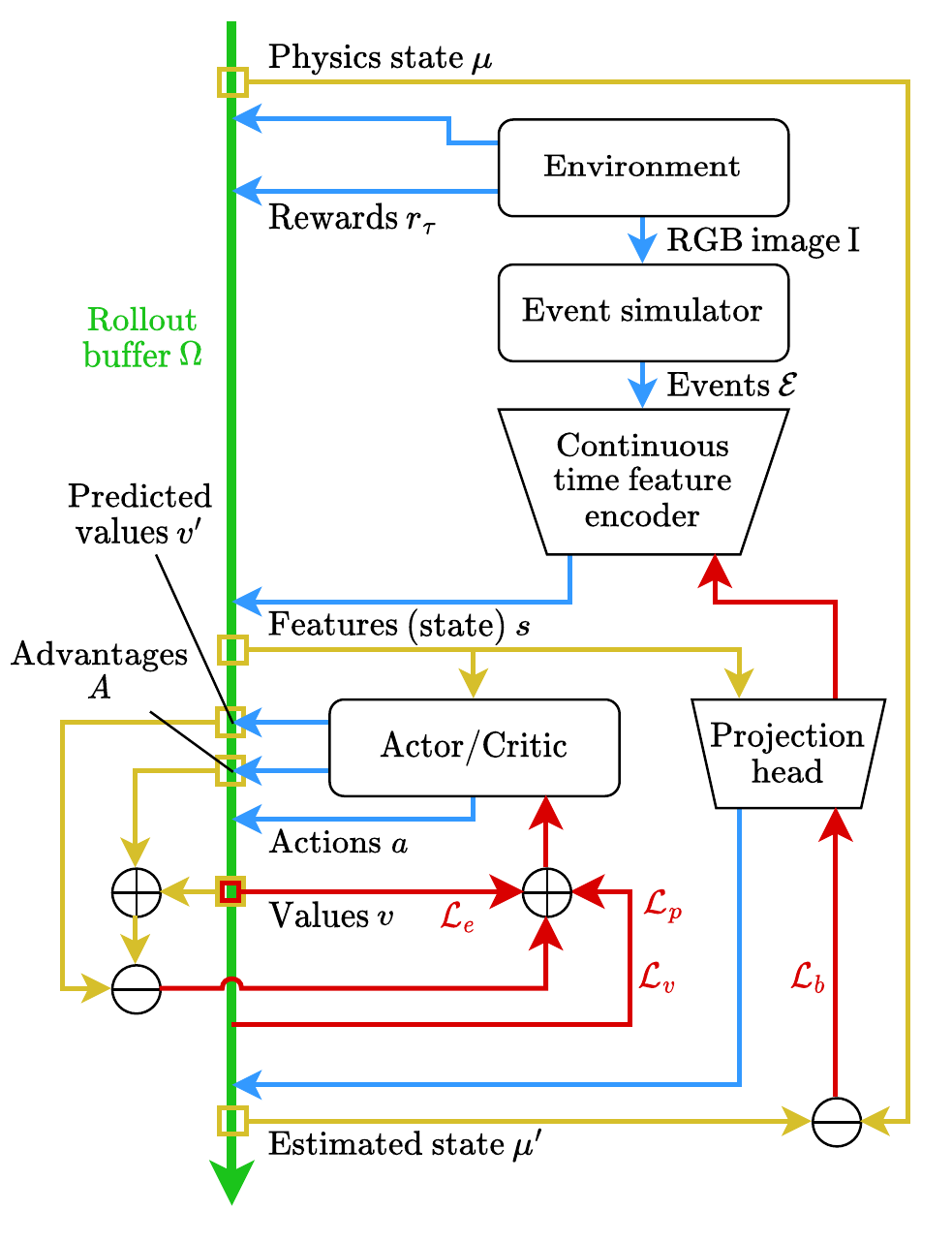}%
	\caption{\label{fig:flowdiagram}
		Flow diagram for training the proposed continuous \acs{RL} system.
		The rollout buffer, shown in {\color{flowGreen}green}, can be appended and sampled over a variable time period.
		{\color{flowBlue}Blue arrows} show the generation of the continuous time rollout (see \cref{sec:rollouts}).
		{\color{flowYellow}Yellow arrows} show how the rollout is sampled.
		{\color{flowRed}Red arrows} show the gradient paths for the policy loss {\color{flowRed}\(\mathcal{L}_p\)}, critic loss {\color{flowRed}\(\mathcal{L}_c\)}, entropy loss {\color{flowRed}\(\mathcal{L}_e\)} and the bootstrap loss {\color{flowRed}\(\mathcal{L}_b\)}.
		Each \(\ominus\) represents the application of the relevant loss function.
	}
\end{figure}

Given this formulation of the continuous time \ac{RL} problem, we propose the \ac{CERiL} solution illustrated in \cref{fig:flowdiagram}.
Here the environment provides frequent but discrete RGB image observations and rewards.
These are then converted into continuous observation streams, and encoded via the continuous feature encoder network.
These feature encodings are finally passed to three network heads.
The projection head attempts to recover the underlying physical state of the system, and is used as a regularisation to help constrain the feature encoder.
The other two heads represent the actor and critic networks of the RL system.
The various subsections of the flow diagram will now be described in turn.

\subsection{Continuous time rollout generation}\label{sec:rollouts}
Although the definitions above generalise to any continuous time \ac{RL} problem, this paper specifically concerns itself with learning from the event stream produced by an event camera.
To generate these continuous time event camera rollouts, we propose a general purpose approach based on the OpenAI Gym~\cite{Brockman2016} environment specification.
Firstly, where possible we modify the environment to reduce the time elapsed per step to \(\frac{1}{N}\).
The parameter \(N\) roughly relates to the density of the event rollout with respect to the control loop.
Next, after each query of the policy network \(\pi(a | \Omega, t)\) we undertake \(N\) repeated steps of the environment using the selected action.
At each step, we record the underlying state vector \(\mu_t\) and render the environment to an RGB image \(I_t\).
We note that both \(\mu_t\) and \(I_t\) are still sampled at regular discrete timesteps (albeit at a greatly increased `framerate').

Next, we pass the image stream \(I\) through the event camera simulator ESIM~\cite{Rebecq2018}.
This produces an event stream \(\mathcal{E} = \{(x, y, p, t)\}\) which is a set of event tuples, each specifying a pixel location \(x, y\), polarity \(p\), and a timestamp \(t\).
These timestamps are no longer discretely spaced, and may occupy any value along the continuous time domain.
This is possible due to the simulator's modelling of event camera behaviour coupled with an interpolation of the brightness values between frames.

We can therefore populate the continuous time rollout via
\begin{equation}\label{eq:event_encoding}
	\Omega(x, y, p, t) = [(x, y, p, t) \in \mathcal{E}]
\end{equation}
where \([\;]\) indicates the Iverson bracket.
This procedure can in principle be followed to create a continuous time event rollout for any OpenAI Gym environment which implements a render function.

\subsection{Continuous time feature encoding}
We manage the sparsity of the continuous event volume by building a feature encoder network using \ac{EDeC} layers~\cite{Walters2022}.

This is superficially similar to performing a 3D convolution on the space-time event volume.
However, unlike a traditional 3D convolution kernel, each \ac{EDeC} kernel includes only a set of \(K \in \mathbb{R}^{n \times n}\) spatial convolution parameters as well as a single temporal decay parameter \(\alpha \in[-1 .. 1]\).
The kernel is defined according to these learnable parameters as
\begin{equation}
	K_p(x, y, t) = K_p(x, y) \alpha^{z - t}
	,
\end{equation}
where \(z\) is the temporal extent of the 3D kernel.
This special structure means the number of learnable parameters per kernel is \(n^2 + 1\) as opposed to \(zn^2\).
More importantly this ensures a densification of the sparse event volume, insuring information propagates across time.
\ac{EDeC} kernels also have a separable structure, which enables extremely efficient online `streaming' inference.

At layer \(l\) of the continuous feature encoder, the encoded rollout for the following layer \(l + 1\) is produced according to
\begin{equation}
	\Omega^{l + 1}(x, y, \bar{p}, t) = \int_{t = 0}^T \sum_p K_p^{\bar{p}} * \Omega^l(x, y, p, t)\alpha^{z - t}
	.
\end{equation}
The input to the first layer of the feature network (\(\Omega^0\)) is the event encoding defined in \cref{eq:event_encoding}.
The final encoded volume which is to be passed to the other network heads, is defined as \(\hat{\Omega}(x, y, p, t)\).

\subsection{State space projection}
Despite the specialised structure of the feature encoding network, it is still challenging to extract useful information directly from the sparse event stream.
In order to help with this, a projection head \(P\) is introduced as a regularisation.
This projection head attempts to recover the stream of underlying physical states (\(\mu\) from \cref{sec:rollouts}) of the environment based on the encoded features.
The corresponding loss is defined as
\begin{equation}
	\mathcal{L}_b(\hat{\Omega}) = \int | P(\hat{\Omega}, t) - \mu_t | dt
	.
	\label{eq:bootstrap}
\end{equation}

It is worth noting that we do not force the encoded feature space to match the underlying physical state.
We only require that the physical state is recoverable.
In reality the encoded features are much higher dimensional and far more expressive than the physics state.
During experimentation we noted that using the ground truth physics state directly (as in many simple fully observed \ac{RL} problems) often leads to reduced performance, when compared to our continuous feature encoding.

\subsection{Actor \& critic decoders}
At the start of \cref{sec:methodology} we defined the continuous time variants of the return, value function and advantage function.
In \ac{CERiL} we use a value network \(V^\pi(\hat{\Omega}, t)\) which estimates the value function at a particular time, given the encoded rollout as input.
The critic loss for this network is defined as
\begin{equation}
	\mathcal{L}_c(\hat{\Omega}) = \int \left| V^\pi(\hat{\Omega}, t) - R(\hat{\Omega}, t, \infty) \right| dt
	.
\end{equation}
We note that although this is defined in the continuous domain, it may only be backpropagated to update the network weights at certain times, dependent on the frequency of the actor.

Our continuous actor network is similarly defined as \(\pi(a | \hat{\Omega}, t)\) which uses the encoded rollout to compute the selection probability of each action, across time.
In order to define the action loss, we first specify the scaling factor \(c\) as
\begin{equation}
	c = \frac{\hat{\pi}(a | \hat{\Omega}, t)}{\pi(a | \hat{\Omega}, t)}
\end{equation}
where \(\hat{\pi}\) is the delayed policy from before the last network update.

This scaling factor allows us to define the loss function as the scaled and clipped version of the continuous advantage function
\begin{equation}
	\mathcal{L}_p(\hat{\Omega}) = \!\int\!\max\left(cA(\hat{\Omega}, t, \hat{a}), clip \left(c, 1 \!-\! \epsilon, 1\! +\! \epsilon\right) A(\hat{\Omega}, t, \hat{a})\right)
\end{equation}
where \(\epsilon\) is a hyperparameter controlling the level of update clipping.
This keeps the update within the trust region.

Finally, an entropy loss is introduced as
\begin{equation}
	\mathcal{L}_e(\hat{\Omega}) = \int \mathbb{E} (-log(
	\pi(a | \hat{\Omega}, t)))
	.
\end{equation}
This ensures that the policy retains an element of exploration.

In practice, the actor, critic and entropy losses are only used to constrain their own respective network heads.
The feature encoding volume itself is primarily constrained by the loss defined in \cref{eq:bootstrap}.

\section{Evaluation}\label{sec:evaluation}

\subsection{Environments}
Four environments with contrasting reward schemes were chosen.
The first environment was \textbf{Pendulum}.
The task is to swing an arm to balance it upright with zero torque.
The physics state is \(\mu = [x, y, \dot{\theta}]\), where \(\theta\) is the orientation of the pendulum, \(x=\cos (\theta)\) and \(y=\sin (\theta)\).
It has a dense reward scheme;
at each step, the reward is a function of the pendulum's angle and the torque,
\begin{equation}
	r = -(\theta^2 + 0.1 * \theta_{\textnormal{dt}}^2 + 0.001 * \textnormal{torque}^2)
	,
\end{equation}
where \(\theta = 0\) is the upright and target position.
The minimum reward is \num{-16.3} and the maximum is zero.

\begin{figure}[!h]\centering
	\includegraphics[width=0.49\linewidth]{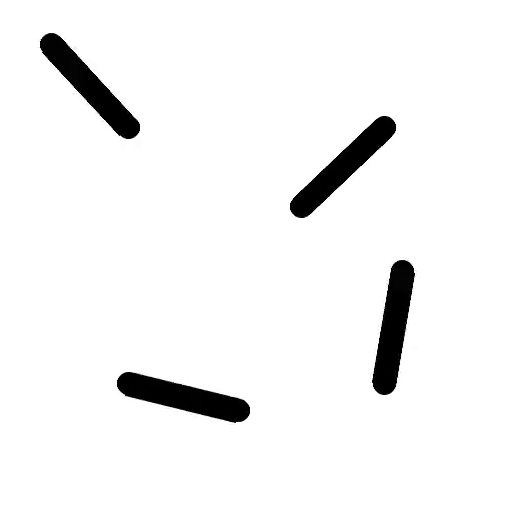}%
	\includegraphics[width=0.49\linewidth]{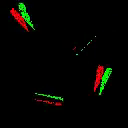}%
	\caption{\label{fig:pendulum}
		Pendulum environment step visualisations (RGB, events)
	}
\end{figure}

For the second environment, \textbf{CartPole}, the agent is a `cart' which can be moved left or right.
A freely-rotating pole is anchored to the agent.
Each step gives a reward of \(+ 1\), and the episode is terminated if the pole falls or if the cart moves too far from the starting position.
The agent should prolong the episode by balancing the pole, reaching a reward cap of \num{500}.
At each timestep, the physics state of this environment is \(\mu = [x, \dot{x}, \theta, \dot{\theta}]\), where \(x\) is the horizontal position, \(\theta\) is the angle of the pole, and the dotted versions are the derivative of their counterparts.

\begin{figure}[!h]\centering
	\includegraphics[width=\linewidth]{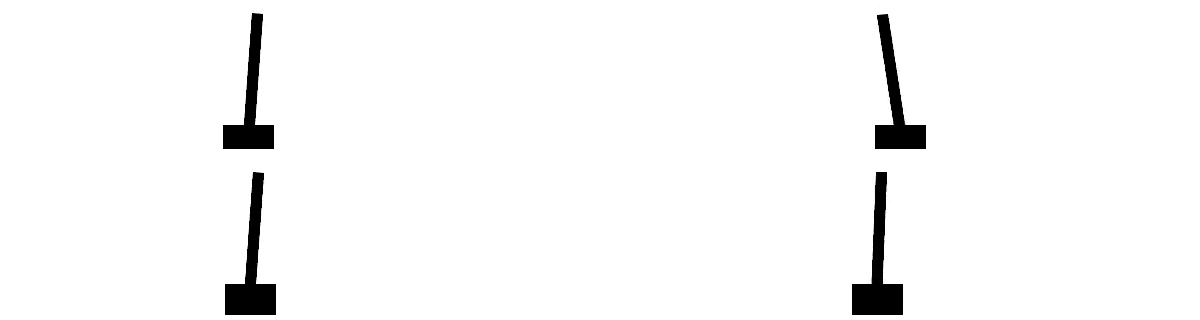}
	\includegraphics[width=\linewidth]{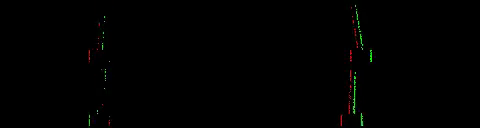}%
	\caption{\label{fig:cartpole}
		CartPole environment step visualisations (RGB, events)
	}
\end{figure}

\textbf{Atari Pong} was chosen for the third environment.
It has a sparse reward scheme;
A reward of +\num{1} or \num{-1} is given when either the agent or the opponent scores a point, respectively. 
Environments with sparse reward schemes are particularly difficult for on-policy algorithms as there is no feedback until something significant happens.
For this environment the physics state \(\mu\) is the RAM state of the atari environment.

\begin{figure}[!h]\centering
	\includegraphics[height=5cm, width=0.49\linewidth]{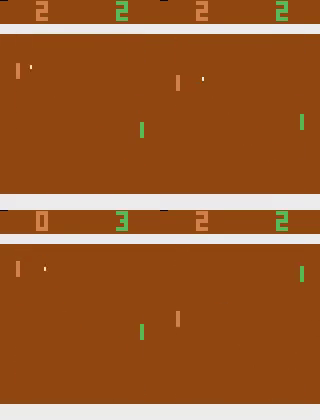}%
	\includegraphics[height=5cm, width=0.49\linewidth]{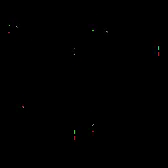}%
	\caption{\label{fig:pong}
		Pong environment step visualisations (RGB, events)
	}
\end{figure}

\textbf{MountainCar} was chosen as the final environment. The physics state of the environment is \(\mu = [x, \dot{x}]\) where \(x\) is the horizontal position of the cart. It has a simple reward scheme which is the opposite of CartPole. A penalty of \num{-1} is applied at every timestep, encouraging early termination of the episode. The only way to terminate the episode is to reach the flag on the right-hand side, which requires a prolonged chain of cycling between the two mountains. This makes the environment incredibly challenging to solve without using imitation learning. Without guidance, it is very rare for random motions to be consistent enough for the cart to reach the goal. The reward signal also does not provide any guidance that the agent is getting closer.

\begin{figure}[!h]\centering
	\includegraphics[width=0.49\linewidth]{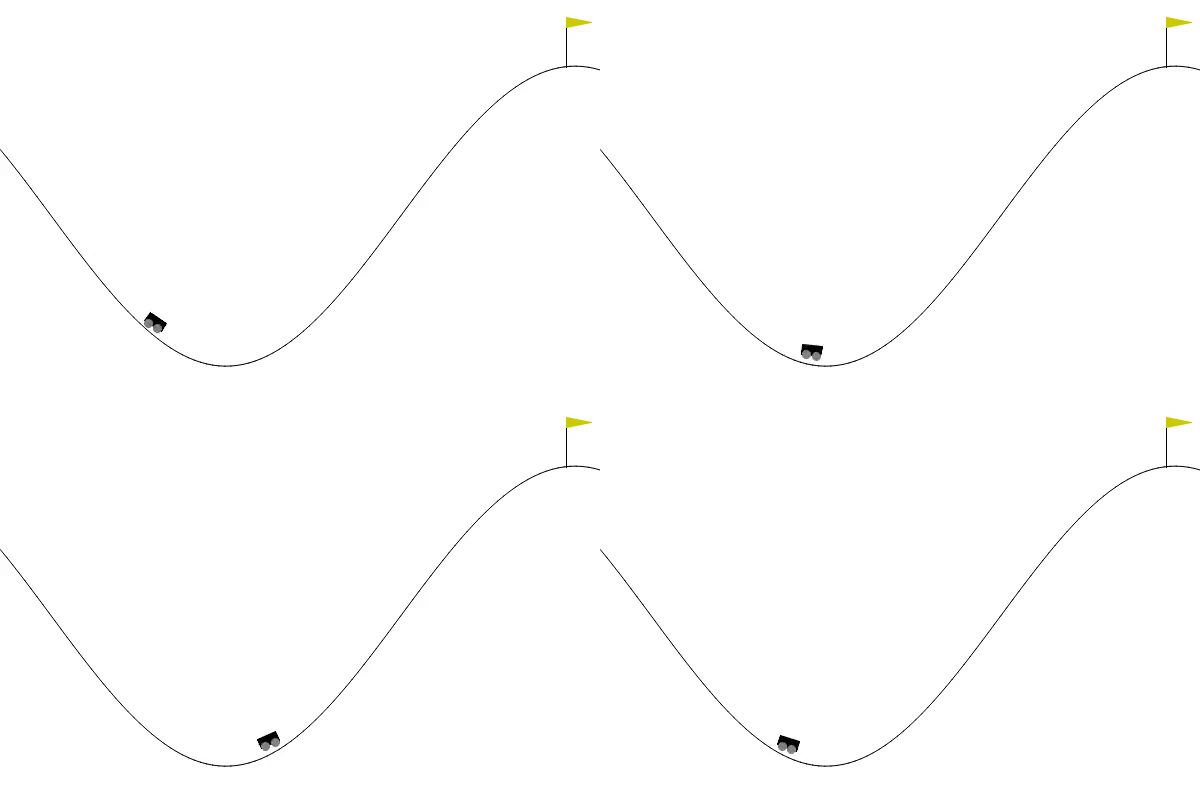}%
	\includegraphics[width=0.49\linewidth]{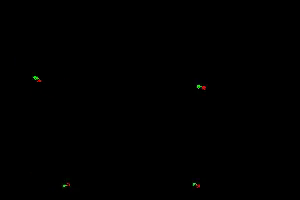}%
	\caption{\label{fig:mountaincar}
		MountainCar environment step visualisations (RGB, events)
	}
\end{figure}

For the event stream observations (shown in \cref{fig:pendulum,fig:cartpole,fig:pong,fig:mountaincar}), note that events are only generated where the brightness value of a given pixel changes.
This means that it is common for some parts of the scene to not be visible to the event camera.
For example, when CartPole's pole is perfectly balanced, it may not generate events.
Additionally, MountainCar's hill and goal are stationary and are therefore not visible.
Nevertheless, the key aspects of these environments are visible through the event camera simulator.
Certain environments like Chess may not be amenable to this approach, as the non-moving pieces are static and thus invisible.
However, the simple introduction of some camera jitter may be enough to circumvent this.

\begin{table*}[h]
	\centering
	\footnotesize
	\begin{tabular}{
		l
		c
		S[table-auto-round,table-format=1.1] 
		S[table-auto-round,table-format=1.1] 
		S[table-auto-round,table-format=1.1] 
		S[table-auto-round,table-format=1.1] 
	}
		\toprule
		Approach & Input data format & \multicolumn{4}{c}{Average rewards}\\
		& & \multicolumn{1}{c}{Pendulum} & \multicolumn{1}{c}{CartPole}  & \multicolumn{1}{c}{Pong} & \multicolumn{1}{c}{MountainCar}\\
		\midrule
		NatureCNN~\cite{Mnih2015} & RGB & -1242.219 & 9.4 & \textbf{17.9}\phantom{0} & -200.0\\
		NatureCNN-e~\cite{Mnih2015} & 2D event image & -1236.561 & 137.4 & 15.04 & -200.0\\
		SNN~\cite{Gehrig2020} & Event stream & -1177.086 & 87.3 & -17.2 & -200.0\\
		\acs{CERiL} (ours) & Event stream & \B{-638.7}\phantom{00.} & \B{438.8}\phantom{0..} & 1.0 & \B{-97.6}\phantom{0.}\\
		\bottomrule
	\end{tabular}
	\caption{\label{tab:results}
		Best average rewards for OpenAI environments.
	}
\end{table*}
\subsection{Baselines}
We train a number of different policy types on each environment.
The first is \textbf{NatureCNN}, a 3-layer \ac{CNN}~\cite{Mnih2015}.
It has been used to achieve human-level control in Atari games with RGB frames as input, and is commonly referred to as the `Nature CNN'.
This is applied to the RGB frames \(I\).
This is disadvantageous compared to state vectors and event streams, since each RGB image does not encode any temporal information.
For example, in MountainCar, it is impossible to tell if the car is rolling up or down the hill, and it is therefore more difficult to build momentum.

The second baseline is \textbf{NatureCNN-e}.
This uses same \ac{CNN} network applied to the event camera data.
To allow the \ac{CNN} to operate on an event stream, the events must be formatted into an image tensor.
To achieve this, we accumulate events within an interval at each pixel location.
We keep the positive and negative events as separate channels, \(C = 2\), and produce aggregated `event frames' as
\begin{equation}
	I^E_t(x, y, p) = \sum_{t}^{t + 1}[(x, y, p, t) \in \mathcal{E}]
	.
\end{equation}
where \([\;]\) indicates the Iverson bracket.

The third baseline is an \textbf{\ac{SNN}}, for which an event stream is the ideal data representation.
This baseline takes a segment of the event stream directly as input, much like the proposed \acs{CERiL} approach.
We use the network architecture proposed in~\cite{Gehrig2020}, with the final layers adapted to match the action space of the environment in question.

\subsection{Results}
We contrast the accuracy of the proposed \acs{CERiL} algorithm against the baselines detailed above.
Every entry for each environment was run with the same network structure (where possible) and hyperparameters.
The actor and critic networks had an architecture consisting of two fully-connected layers with \num{64} units, with Tanh non-linearity functions applied after each.

\begin{figure}[h]\centering
	\includegraphics[width=\linewidth]{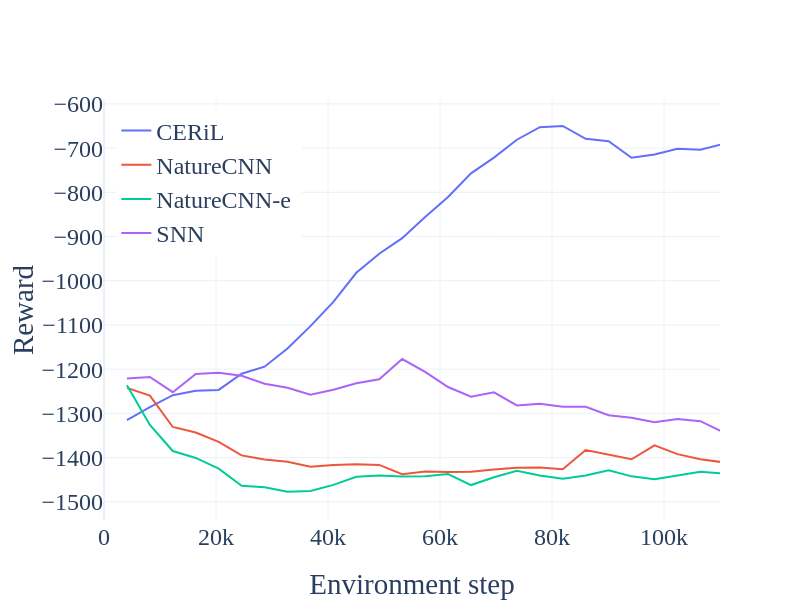}%
	\caption{\label{fig:pendulum2}
		Average rewards for Pendulum in evaluation environment.
	}
\end{figure}

\begin{figure}[h]\centering
	\includegraphics[width=\linewidth]{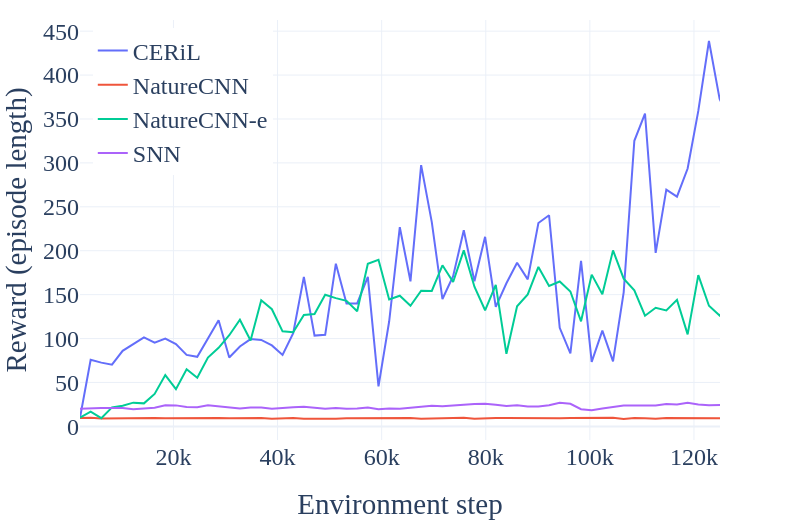}%
	\caption{\label{fig:cartpole2}
		Average rewards for CartPole in evaluation environment.
	}
\end{figure}

\begin{figure}[h]\centering
	\includegraphics[width=\linewidth]{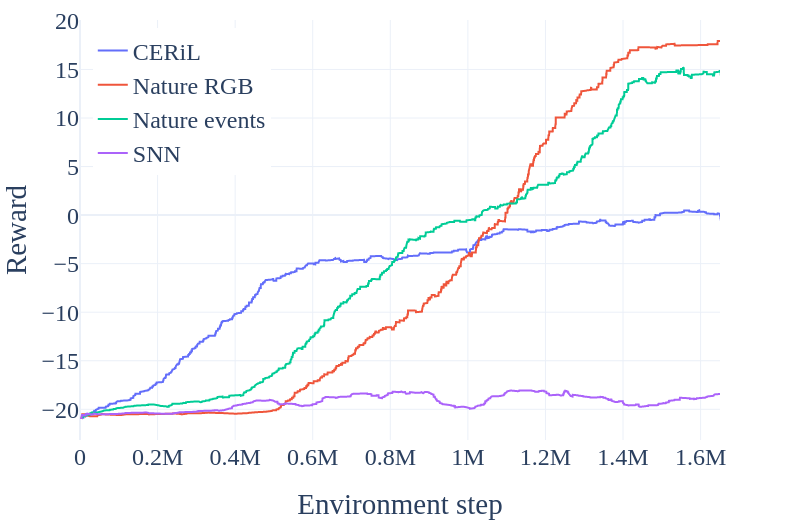}%
	\caption{\label{fig:pong2}
		Average rewards for Pong in evaluation environment.
	}
\end{figure}

\begin{figure}[h]\centering
	\includegraphics[width=\linewidth]{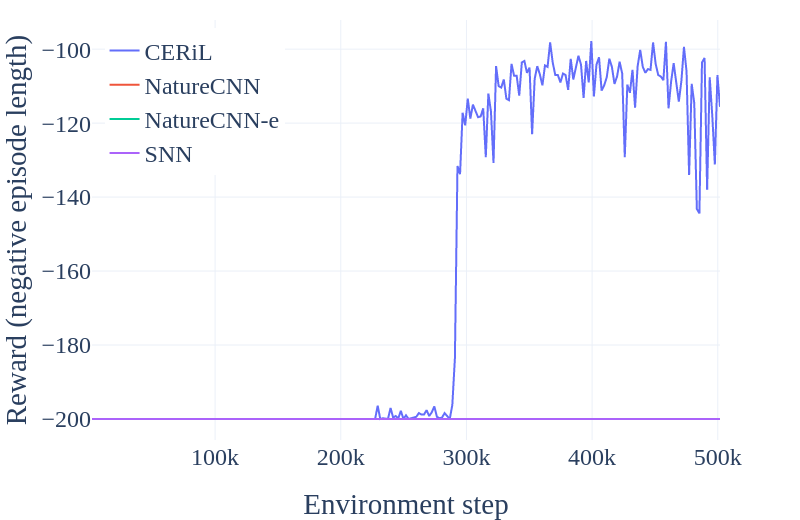}%
	\caption{\label{fig:mountaincar2}
		Average rewards for MountainCar in evaluation environment.
	}
\end{figure}

The results for each environment are shown in \cref{tab:results} and \cref{fig:pendulum2,fig:cartpole2,fig:pong2,fig:mountaincar2}.

For the first environment, Pendulum, the proposed \acs{CERiL} approach clearly outperformed the others.
When examining the videos, \acs{CERiL} manages to balance the pendulum upright for short amounts of time.
However, during these times, the event stream is almost empty.
The data-propagating nature of \acs{CERiL} could be the main factor that sets it apart from the other event-based approaches, as events from previous frames (on the swing-up) provide a signal beyond what is immediately visible.
The best reward achieved did reach the `completion' state of zero reward for at least one episode in each technique.
This could indicate that the period of time balancing the pendulum without receiving events may be too long, and perhaps a more sensitive event generator could allow the event-based \ac{RL} networks to approach optimal performance.

For the second environment, CartPole, the NatureCNN operating on RGB frames was not able to solve the task.
This is likely because there is no temporal information encoded in the image frames.
As a result, it is not possible to determine if the pole is falling.
Note that it is possible to observe changes over time through frame-stacking, although the effective time period would end up being many times longer than a single step.
The NatureCNN operating on an aggregated event images fared better.
However, it may sometimes be difficult to determine the direction of motion from event images.
Our \acs{CERiL} approach was able to achieve a high reward averaging close to the upper threshold, as it can leverage events in their original order.

For the third environment, Pong, the proposed \acs{CERiL} approach begins learning much more rapidly than all other approaches.
At 0.6M steps, the performance is greater than that of all competing approaches combined.
However, at this point learning slows dramatically and is overtaken by the. NatureCNN approaches at around 1M steps.
The \ac{SNN} baseline proves unable to make any significant progress in this environment.
It is likely that there are some intricacies of the environment that are hard to capture with the shorter timescales of the event stream inputs.
For example, the fact that when the opponent's paddle remains stationary it disappears from the observation space.
The longer-term aggregation of the NatureCNN approaches may mitigate this.

For the final environment, MountainCar, a solution threshold of \num{-110} is typically set, which when achieved, the model is considered to have solved the environment.
Neither the NatureCNN running on the rendered image nor the event-based NatureCNN-e on the event stream were able to breach \num{-200} reward.
This means that the car never made it to the goal through random exploration by the time limit, despite many attempts with different seeds.
In contrast, our \acs{CERiL} approach achieved an average reward of \num{-97.6}, which exceeds the completion threshold.
As with CartPole, the network trained on RGB frames cannot resolve velocity, impairing exploration.
Both the aggregated event images and event stream are not very informative at episode starts because the car has little momentum and moves slowly.
Nevertheless, utilising a continuous stream enables the environment to be solved.

\section{Conclusions}\label{sec:conclusion}
In this paper we have proposed \acl{CERiL}.
This is a novel approach to \ac{RL} where a continuous stream of input observations leads to a continuous sequence of output actions.
As such, the environmental updates and reward feedback may be asynchronous and at irregular intervals.
The resulting system is able to be far more reactive than traditional visual \ac{RL} systems.
Furthermore, we propose a framework by which any traditional \ac{RL} environment can be made continuous through the use of the event camera simulator.
In order to produce an observation stream for \acs{CERiL} in the real world, we propose to use an event-camera sensor.

Our results show the promise of event streams in \ac{RL}, which the proposed system utilises to outperform previous benchmarks.
The code for this framework will be released to the community in order to verify the results and support further research in this area.

As future work, it would be interesting to explore the implication of continuous time \ac{RL} more broadly.
The idea could be extended to other \ac{RL} algorithms besides \ac{PPO}, such as A3C, DDPG, or SAC.
It may also be useful to explore the implications for related areas such as inverse \ac{RL} or imitation learning.
Finally, the results on the Pong environment imply that there may be some benefit from a combined visual-event framework, which can maintain information about stationary environmental objects.

\bibliography{bibliography}
\bibliographystyle{icml2023}

\end{document}